\documentclass{article}

\usepackage[preprint]{neurips_2026}

\usepackage[utf8]{inputenc}
\usepackage[T1]{fontenc}
\usepackage{hyperref}
\usepackage{url}
\usepackage{booktabs}
\usepackage{array}
\usepackage{amsfonts}
\usepackage{amsmath,amssymb}
\usepackage{nicefrac}
\usepackage{microtype}
\usepackage{xcolor}
\usepackage{graphicx}
\usepackage{tikz}
\usetikzlibrary{positioning, arrows.meta, fit, backgrounds, shapes.geometric, calc}
\usepackage{algorithm}
\usepackage{algpseudocode}

\graphicspath{{figures/}}
\title{Causal Reinforcement Learning for Complex Card Games: A \textit{Magic The Gathering} Benchmark}

\author{Cristiano da Costa Cunha, Wei Liu, Tim French, and Ajmal Mian
\thanks{C. da Costa Cunha, W. Liu, T. French, and A. Mian are with the Department of Computer Science and Software Engineering, University of Western Australia, 6009 WA Australia (e-mail: cris.dacostacunha@research.uwa.edu.au; wei.liu@uwa.edu.au; tim.french@uwa.edu.au; ajmal.mian@uwa.edu.au).}}

\begin{document}

\maketitle

\begin{abstract}
    Causal reinforcement learning (RL) lacks benchmarks for complex systems 
    that combine sequential decision making, hidden information, large masked action spaces, and explicit causal structure. We introduce
    \textbf{MTG-Causal-RL}, a Gymnasium benchmark built on \textit{Magic:
    The Gathering} with a 3{,}077-dimensional partial observation, a
    478-action masked discrete action space, five competitive Standard
    archetypes, three reward schemes, and a hand-specified Structural
    Causal Model (SCM) over strategic variables. Every episode exposes
    causal variables, SCM-predicted intervention effects, and per-factor
    credit traces, making causal credit assignment, leave-one-out
    cross-archetype transfer, and policy auditability first-class
    benchmark metrics. We adapt a panel of reference baselines (random,
    heuristic, masked PPO, a causal-world-model PPO variant, an
    architecture-matched scalar control) and propose Causal Graph-Factored
    Advantage PPO (CGFA-PPO) as a secondary contribution: a causal agent
    that uses the SCM parents of win probability as factor-aligned critic
    targets together with an intervention-calibration loss. All
    comparisons use paired seeds, paired-bootstrap confidence intervals, and
    Holm-Bonferroni correction within pre-registered families. Masked
    PPO and CGFA-PPO reach competitive in-distribution win rates and
    clearly exceed the random baseline; per-factor calibration
    trajectories and leave-one-out transfer gaps then expose diagnostic
    structure that scalar win rate alone cannot. We
    release the benchmark, the reference-baseline results, and the full
    evaluation protocol openly at
\url{https://anonymous.4open.science/r/mtg-causal-rl-E927}. By coupling a strategically rich,
    partially observed domain with an explicit causal interface and a
    pre-registered statistical protocol, MTG-Causal-RL gives causal-RL,
    world-model, and LLM-agent research a shared testbed for open
    questions current benchmarks cannot pose together: causal credit
    assignment under masked action spaces, structural transfer across
    cleanly defined archetypes, and SCM-grounded policy auditability.
    \end{abstract}

\section{Introduction}
\label{sec:intro}

RL has reached or surpassed expert humans in fully observable
games~\citep{silver2016mastering, silver2017mastering,
silver2018general}, imperfect-information
poker~\citep{brown2018superhuman, brown2019superhuman}, and pixel
domains~\citep{mnih2015human}. Yet \emph{causal RL} still lacks
benchmarks that combine sequential decisions with structural causal
variables, interventions, and transfer tests~\citep{pearl2009causality,
scholkopf2021toward, zeng2023causalrl}. Existing environments offer
either low-level control with clear physical
interventions~\citep{todorov2012mujoco, tassa2018deepmind} or
strategic games without an explicit causal
interface~\citep{bellemare2013arcade, cobbe2020procgen,
kuttler2020nethack}.

\textit{Magic: The Gathering} (MTG)~\citep{garfield1995mtg,
churchill2019magic, wotc2024mtgrules} is partially observable,
stochastic, and combinatorial, with multi-phase turns and masked
actions over \textit{mulligans}, \textit{sequencing}, \textit{spell timing}, \textit{combat}, \textit{targeting},
and \textit{mana payment}. It also has natural causal variables: mana enables
spells, board pressure changes combat incentives, card advantage
changes option value, and life buffer changes risk tolerance.

\textbf{MTG-Causal-RL} turns those properties into a reusable
interface for causal RL research. We provide a
Gymnasium~\citep{towers2024gymnasium} environment with a 478-action
masked discrete action space, a 3{,}077-dimensional partial
observation, five competitive Standard 2025 archetypes, three reward
schemes, and an explicit Structural Causal Model (SCM) over strategic
variables. The SCM is part of the interface: every episode exposes
causal variables, SCM-predicted intervention effects, and per-factor
credit traces. We adapt reference baselines and, as a second
contribution, propose Causal Graph-Factored Advantage PPO (CGFA-PPO):
per-factor critic heads indexed by the SCM parents of
$\mathrm{WinProb}$, a state-conditional gate, and an
intervention-calibration loss.

\paragraph{Contributions and outcomes.}
We contribute (i) the MTG-Causal-RL benchmark environment and causal
interface (primary); (ii) a reference baseline panel adapted to it;
(iii) CGFA-PPO as a reference causal method (secondary); and (iv) a
paired-seed evaluation protocol with Wilson
CIs~\citep{wilson1927probable}, paired-bootstrap
tests~\citep{efron1993bootstrap}, Wilcoxon signed-rank
tests~\citep{wilcoxon1945individual}, and Holm-Bonferroni
correction~\citep{holm1979simple} within pre-registered comparison
families. All code, deck definitions, training and evaluation
manifests, and the SCM specification are released openly at
\url{https://anonymous.4open.science/r/mtg-causal-rl-E927}. PPO and CGFA-PPO
both exceed the uniform random baseline, but the headline comparison
is deck-dependent rather than uniformly favouring the causal agent:
CGFA-PPO improves over PPO on some archetypes and trails it on others.
The leave-one-out cross-archetype transfer gap, ablation suite, and
per-factor calibration trajectory then expose benchmark diagnostics
beyond headline win rate
(Section~\ref{sec:experiments}, Table~\ref{tab:main_results}).

\section{Related Work}
\label{sec:related}

\paragraph{Game benchmarks for RL.}
Deep RL has advanced Atari, board games, and
poker~\citep{mnih2015human, silver2016mastering, silver2017mastering,
silver2018general, brown2018superhuman, brown2019superhuman}.
Hidden-information card games have been studied via ISMCTS,
collectible-card-game AI suites, DouZero, Bridge, and
Hearthstone~\citep{cowling2012ismcts, kowalski2023locm,
zha2021douzero, yeh2016bridge, santos2017monte,
hoover2020hearthstone}. MTG has received less attention, despite
draft-search work~\citep{ward2009mtgdraft}, partial observability, and
formal computational complexity~\citep{churchill2019magic};
MTG-Causal-RL adds an explicit causal abstraction to this family.

\paragraph{Causal RL and existing benchmarks.}
Causal RL connects structural causal
modelling~\citep{pearl2009causality, pearl2019seven} with sequential
decision making~\citep{zeng2023causalrl, scholkopf2021toward},
including causal bandits, counterfactual policy search, causal
dynamics, invariant transfer, and causal world
models~\citep{bareinboim2015bandits, lu2018deconfounding,
zhang2020imitation, buesing2018woulda, forney2017counterfactual,
wang2022causaldynamics, zhang2020invariant, huang2022adarl,
richens2024robust}. CausalWorld~\citep{ahmed2020causalworld} and
MAGICAL~\citep{toyer2020magical} emphasise manipulation and imitation. Real-world applications of Causal RL Causal are high-stakes sequential decisions in healthcare, robotics, autonomous systems, finance, and operations by learning not just which actions work, but why they work under interventions and changing conditions~\citep{scholkopf2021toward,zeng2023causalrl}.
MTG-Causal-RL targets strategic play with masked actions, hidden
information, stochastic draws, and opponent-conditioned strategy.

\paragraph{Reference causal-agent architectures.}
World models and planners such as Dreamer, DreamerV2, and
MuZero~\citep{hafner2020dreamer, hafner2021mastering,
schrittwieser2020muzero} learn predictive models from interaction;
UNREAL and Horde add auxiliary prediction~\citep{jaderberg2017unreal,
sutton2011horde}; HRA and Option-Critic factor reward or
behaviour~\citep{vanseijen2017hra, bacon2017option}. Our
causal-world-model baseline exercises auxiliary causal prediction,
while CGFA-PPO exercises structured advantages through SCM-factor
critics, a gate, and calibration loss.

\paragraph{Generalisation, transfer, and statistics for causal claims.}
Causal claims require held-out and interventional tests, not only
training-distribution win rates. MTG-Causal-RL therefore bundles
seed-paired comparisons, bootstrap and rank-based tests, and
family-level correction~\citep{henderson2018matters, agarwal2021deep,
cobbe2019quantifying, kirk2023survey, wilson1927probable,
efron1993bootstrap, welch1947generalization, wilcoxon1945individual,
hodges1963lehmann, holm1979simple}. Its Gymnasium interface also lets
LLM-based strategic agents plug in directly~\citep{costarelli2024gamebench,
duan2024gtbench}.

\section{The MTG-Causal-RL Environment}
\label{sec:env}

\paragraph{Game overview.}
\textit{Magic: The Gathering}~\citep{garfield1995mtg,
churchill2019magic, wotc2024mtgrules} is a two-player collectible card
game with over 27{,}000 cards. Players start with 20 life and a
60-card deck, draw cards, play mana-producing lands, cast spells, and
usually win by reducing the opponent to zero life. MTG combines mana
sequencing, card advantage, partial observability, stochastic draws,
and formal computational complexity~\citep{churchill2019magic},
motivating learned policies and approximate evaluation.

\paragraph{Task and environment.}
We focus on Standard-format decks~\citep{wotc2024standard,
wizards2024metagame, mtggoldfish2024meta}. Episodes end at lethal
damage or a user-selected turn cap. Table~\ref{tab:benchmark_config}
summarises the Gymnasium~\citep{towers2024gymnasium} environment,
which integrates with Stable-Baselines3~\citep{raffin2021stable}. The
observation is a 3{,}077-dimensional vector over visible hand,
battlefield, graveyard, life, phase, and mana state; the 478-action
space is organised into 16 masked categories~\citep{huang2022masking}.
Only 2 to 15 actions are typically legal, and each MTG turn requires
10 to 20 sequential agent decisions.
Appendix~\ref{app:actions} lists the exact action index ranges.

\begin{table}[t]
\centering
\caption{MTG-Causal-RL benchmark configuration.}
\label{tab:benchmark_config}
\small
\begin{tabular}{@{}>{\raggedright\arraybackslash}p{0.22\linewidth}>{\raggedright\arraybackslash}p{0.72\linewidth}@{}}
\toprule
Component & Specification \\
\midrule
Card pool & 56 unique cards (Standard 2025 legal) \\
Deck size & 60; 5 archetypes (Mono-Red Aggro, Azorius Control,
                Dimir Midrange, Domain Ramp, Boros Convoke) \\
Turn cap & User selectable; benchmark recipes report the chosen value \\
Starting life & 20; starting hand 7 cards \\
Observation & Partial; 3{,}077-dimensional flat vector \\
Action space & 478 masked discrete actions across 16 categories \\
Reproducibility & Deterministic given seed; manifest captures git,
                lockfile, and runtime \\
\bottomrule
\end{tabular}
\end{table}

\paragraph{Reward schemes.}
We provide \textit{sparse} terminal reward, \textit{shaped}
potential-based reward over agent-observable causal variables, and a
\textit{dense} variant adding per-step damage, card-draw, and creature
signals. The shaped reward
$R^{\mathrm{shaped}}_t = \gamma\,\Phi(s_{t+1}) - \Phi(s_t)
+ R^{\mathrm{terminal}}_t$ follows
potential-based shaping under partial
observability~\citep{ng1999policy, eck2016potential};
Appendix~\ref{app:reward_shaping} gives the full derivation.

\section{Causal Abstraction}
\label{sec:causal}

\begin{figure}[t]
\centering
\resizebox{\textwidth}{!}{%
\begin{tikzpicture}[
    node distance=10mm and 12mm,
    every node/.style={font=\small},
    res/.style={draw, fill=blue!10, rounded corners=1mm, inner sep=2pt,
                minimum height=6mm, align=center},
    bs/.style={draw, fill=orange!15, rounded corners=1mm, inner sep=2pt,
                minimum height=6mm, align=center},
    sp/.style={draw, fill=green!15, rounded corners=1mm, inner sep=2pt,
                minimum height=6mm, align=center},
    outcome/.style={draw, fill=red!15, rounded corners=1mm, inner sep=2pt,
                minimum height=6mm, align=center, font=\small\bfseries},
    arr/.style={-{Stealth[length=1.5mm]}, thin, gray!70},
]
    \node[res] (mana_t) {$\mathrm{Mana}_t$};
    \node[res, right=of mana_t] (mana_creat) {$\mathrm{ManaCreat.}$};
    \node[res, right=of mana_creat] (land_drop) {$\mathrm{LandDrop}$};
    \node[res, right=of land_drop] (card_count) {$\mathrm{CardCount}$};
    \node[res, right=of card_count] (has_rem) {$\mathrm{HasRemoval}$};

    \node[res, below=8mm of mana_creat] (mana_t1) {$\mathrm{Mana}_{t+1}$};

    \node[bs, below=12mm of mana_t1] (board_press) {$\mathrm{BoardPress}$};
    \node[bs, right=12mm of board_press] (threat_dens) {$\mathrm{ThreatDensity}$};

    \node[sp, below=12mm of board_press, xshift=-15mm] (card_adv) {$\mathrm{CardAdv}$};
    \node[sp, right=6mm of card_adv] (tempo) {$\mathrm{Tempo}$};
    \node[sp, right=6mm of tempo] (life_buf) {$\mathrm{LifeBuffer}$};
    \node[sp, right=6mm of life_buf] (rem_avail) {$\mathrm{RemovalAvail}$};

    \node[outcome, below=12mm of tempo, xshift=8mm] (win) {$\mathrm{WinProb}$};

    \draw[arr] (mana_t)     -- (mana_t1);
    \draw[arr] (mana_creat) -- (mana_t1);
    \draw[arr] (land_drop)  -- (mana_t1);

    \draw[arr] (mana_t)     -- (tempo);
    \draw[arr] (mana_t1)    -- (board_press);
    \draw[arr] (mana_t1)    -- (threat_dens);

    \draw[arr] (board_press) -- (card_adv);
    \draw[arr] (board_press) -- (tempo);
    \draw[arr] (board_press) -- (win);

    \draw[arr] (threat_dens) -- (win);
    \draw[arr] (threat_dens) -- (board_press);

    \draw[arr] (card_count) -- (card_adv);
    \draw[arr] (has_rem)    -- (rem_avail);
    \draw[arr] (rem_avail)  -- (win);

    \draw[arr] (card_adv)  -- (win);
    \draw[arr] (tempo)     -- (win);
    \draw[arr] (life_buf)  -- (win);

    \begin{scope}[on background layer]
        \node[draw=blue!50, dashed, rounded corners,
              fit=(mana_t)(mana_creat)(land_drop)(card_count)(has_rem)(mana_t1),
              inner sep=3pt, label={[blue!60, font=\scriptsize]left:Resources}]
              {};
        \node[draw=orange!60, dashed, rounded corners,
              fit=(board_press)(threat_dens),
              inner sep=3pt, label={[orange!70, font=\scriptsize]left:Board state}]
              {};
        \node[draw=green!50!black, dashed, rounded corners,
              fit=(card_adv)(tempo)(life_buf)(rem_avail),
              inner sep=3pt,
              label={[green!50!black, font=\scriptsize]left:Strategic}]
              {};
        \node[draw=red!50, dashed, rounded corners,
              fit=(win), inner sep=4pt,
              label={[red!60, font=\scriptsize]left:Outcome}]
              {};
    \end{scope}
\end{tikzpicture}%
}
\caption{Hand-designed structural causal model for MTG-Causal-RL.
Variables are organised in four layers: resources (blue), board state
(orange), strategic position (green), and outcome (red). Edges denote
causal dependence as enforced by the structural equations
(Appendix~\ref{app:scm_eqs}). The six parents of $\mathrm{WinProb}$
form the benchmark's strategic causal factors.}
\label{fig:scm}
\end{figure}
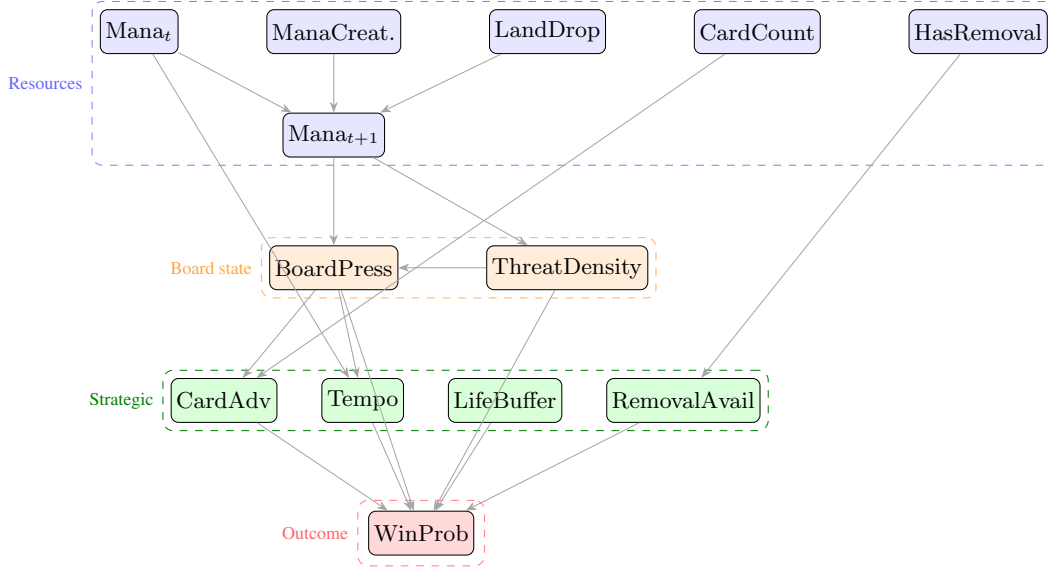

The Structural Causal Model (SCM) is hand-designed from MTG strategic
theory rather than learned from data, prioritising interpretability and
edge orientation~\citep{scholkopf2021toward}. Table~\ref{tab:causal_vars}
lists the variables in four layers (resources, board state, strategic
position, outcome). Figure~\ref{fig:scm} shows the graph;
Appendix~\ref{app:scm_eqs} gives the equations and Pearl-style
$\mathrm{do}(\cdot)$ propagation used to compute intervention effects
$\varepsilon_k(s,a) = \phi_k(s') - \phi_k(s)$ analytically.

\begin{table}[t]
\centering
\caption{Causal variables in the MTG-Causal-RL SCM. Variables are
organised by causal layer; the six parents of $\mathrm{WinProb}$
define the benchmark's strategic causal factors.}
\label{tab:causal_vars}
\small
\begin{tabular}{llp{6.6cm}}
\toprule
Variable & Range & Operational definition \\
\midrule
\multicolumn{3}{l}{\textit{Resource layer}} \\
$\mathrm{Mana}_t$ & 0--10 & Mana-producing permanents controlled \\
$\mathrm{LandDrop}_t$ & $\{0,1\}$ & At least one land in hand \\
$\mathrm{ManaCreatures}$ & 0--10 & Mana-producing creatures on battlefield \\
$\mathrm{Mana}_{t+1}$ & 0--10 &
    $\mathrm{Mana}_t + \mathbf{1}[\mathrm{LandDrop}_t] + \mathrm{ManaCreatures}$ \\
$\mathrm{CardCount}$ & 0--15 & Cards in hand \\
$\mathrm{HasRemoval}$ & $\{0,1\}$ & At least one removal spell in hand \\
\midrule
\multicolumn{3}{l}{\textit{Board state layer}} \\
$\mathrm{BoardPress}$ & $[-20,20]$ & Net creature power (own minus opponent) \\
$\mathrm{ThreatDensity}$ & $[0,1]$ & Threats divided by total own permanents \\
\midrule
\multicolumn{3}{l}{\textit{Strategic position layer}} \\
$\mathrm{CardAdv}$ & $[-10,10]$ & Own permanents minus opponent permanents \\
$\mathrm{Tempo}$ & $[-1,1]$ & Mana efficiency differential, clipped \\
$\mathrm{LifeBuffer}$ & $[-20,20]$ & Own life minus opponent life \\
$\mathrm{RemovalAvail}$ & $\{0,1\}$ & Pass-through of $\mathrm{HasRemoval}$ \\
\midrule
\multicolumn{3}{l}{\textit{Outcome layer}} \\
$\mathrm{WinProb}$ & $(0,1)$ &
    $\sigma\!\big(\mathbf{w}^{\!\top} \phi\big)$ over its six parents \\
\bottomrule
\end{tabular}
\end{table}

\paragraph{Actions as interventions.}
Actions map to interventions through card-level mechanics (removal,
direct damage, card draw, counterspell). Playing a land, for example,
performs $\mathrm{do}(\mathrm{LandDrop}=1)$, which propagates through
$\mathrm{Mana}_{t+1}$ and board variables to $\mathrm{WinProb}$. This
supports post-intervention $\mathrm{WinProb}$ prediction, per-factor
deltas $\varepsilon_k(s,a)$, and causal credit assignment; the shaped
reward uses the same variables as a potential
function~\citep{ng1999policy, eck2016potential}.

\section{Reference Methods}
\label{sec:methods}

\paragraph{Notation.}
We use standard RL notation~\citep{sutton2018reinforcement}: $s_t$,
$a_t$, $r_t$, scalar critic $V(s_t)$, and PPO advantage
$A^{\mathrm{scalar}}(s_t,a_t)$~\citep{schulman2016gae,
schulman2017proximal, huang2022masking}. Let
$\mathrm{CV}_t \in \mathbb{R}^{K}$ be the benchmark causal variables
(default $K=6$, the parents of $\mathrm{WinProb}$),
$\Delta \mathrm{CV}_t$ their per-step change, and
$\varepsilon_{k,t}$ the SCM-predicted intervention effect for factor
$k$.

\paragraph{Baselines.}
\label{sec:baselines}
All baselines share the observation, action, and reward specification.
\textbf{Random} samples legal actions uniformly. \textbf{Heuristic}
uses one rule-based script per archetype. \textbf{PPO} is vanilla
masked PPO via Stable-Baselines3~\citep{raffin2021stable}.
\textbf{Causal Agent (CWM-augmented PPO)} predicts
$\Delta\mathrm{CV}_t$ and a learned win-probability head, then blends
the PPO log-probability with the CWM-predicted change, isolating
auxiliary causal supervision without per-factor advantage
decomposition. \textbf{CGFA scalar-only} matches CGFA-PPO's parameter
count but zeros the CGFA losses and feeds only
$A^{\mathrm{scalar}}$ to the policy update.

\paragraph{CGFA-PPO (reference causal agent).}
\label{sec:cgfa}
CGFA-PPO exercises the causal interface without replacing the
environment or evaluation protocol. Let $\phi(s) \in \mathbb{R}^{K}$
be the benchmark's strategic factors, the parents of
$\mathrm{WinProb}$ in Table~\ref{tab:causal_vars}. CGFA-PPO keeps the
masked PPO actor and scalar critic $V(s)$ and adds three components.
\textit{(i)~A per-factor critic} $V_k(s)$, $k = 1, \dots, K$,
trained against the per-factor reward
$r^{\mathrm{factor}}_{k,t} = \phi_k(s_{t+1}) - \phi_k(s_t)$ using
the same generalised-advantage truncation as the scalar critic,
producing per-factor advantages $A_k(s, a)$.
\textit{(ii)~Learnable mixture weights}
$w_k = \mathrm{softmax}(\beta)_k$ over the $K$ factor heads,
initialised from the SCM's logistic-regression weights so the agent
starts with the structural prior.
\textit{(iii)~A state-conditional residual gate}
$g(s) \in (0, 1)$ that decides per state how much the policy update
should rely on the scalar versus the factor-aligned advantage.
The advantage actually fed to the PPO surrogate is the residual
blend
\[
A_{\mathrm{used}}(s, a) =
\bigl(1 - g(s)\bigr)\, A^{\mathrm{scalar}}(s, a)
+ g(s) \sum_{k=1}^{K} w_k\, A_k(s, a).
\]
A Pearson-correlation calibration loss aligns each factor head
with its SCM-predicted intervention target,
\[
\mathcal{L}_{\mathrm{cal}} =
- \frac{1}{K} \sum_{k=1}^{K}
\frac{\operatorname{Cov}_t(A_{k,t}, \varepsilon_{k,t})}
     {\sigma_{A_k}\, \sigma_{\varepsilon_k} + \delta},
\]
where $\operatorname{Cov}_t$ and $\sigma_{(\cdot)}$ are minibatch
statistics and $\delta>0$ is a stability constant. The benchmark logs
factor-target sign agreement, gate collapse, and factor credit shares;
full losses, architecture, pseudocode, and hyperparameters are in
Appendices~\ref{app:hyperparams} and~\ref{app:cgfa_algorithm}.

\section{Experiments}
\label{sec:experiments}

\paragraph{Protocol.}
Our full protocol uses at least $10^{6}$ environment steps per
opponent, 7 paired seeds per (agent, deck) cell, and 300 deterministic
evaluation episodes per (agent, seed, opponent). Win rates use Wilson
and paired-bootstrap CIs~\citep{wilson1927probable, efron1993bootstrap};
pairwise tests use paired bootstrap, Wilcoxon signed-rank, and
Holm-Bonferroni correction within pre-registered
families~\citep{wilcoxon1945individual, hodges1963lehmann, holm1979simple}
(Appendix~\ref{app:stats}). The submitted reference bundle is a
computationally bounded \emph{holistic} sweep covering all five
archetypes, baselines, ablations, transfer, calibration, and case
studies. The learned-agent headline and transfer comparisons in
Tables~\ref{tab:main_results}--\ref{tab:transfer_gap} use $n_s=2$
paired seeds and 30 episodes per opponent, so their p-values are
exploratory diagnostics rather than confirmatory claims. Each run
writes a reproducibility manifest with git, lockfile, runtime, CLI,
hyperparameters, and seeds.

\begin{figure}[!t]
\centering
\includegraphics[width=\linewidth]{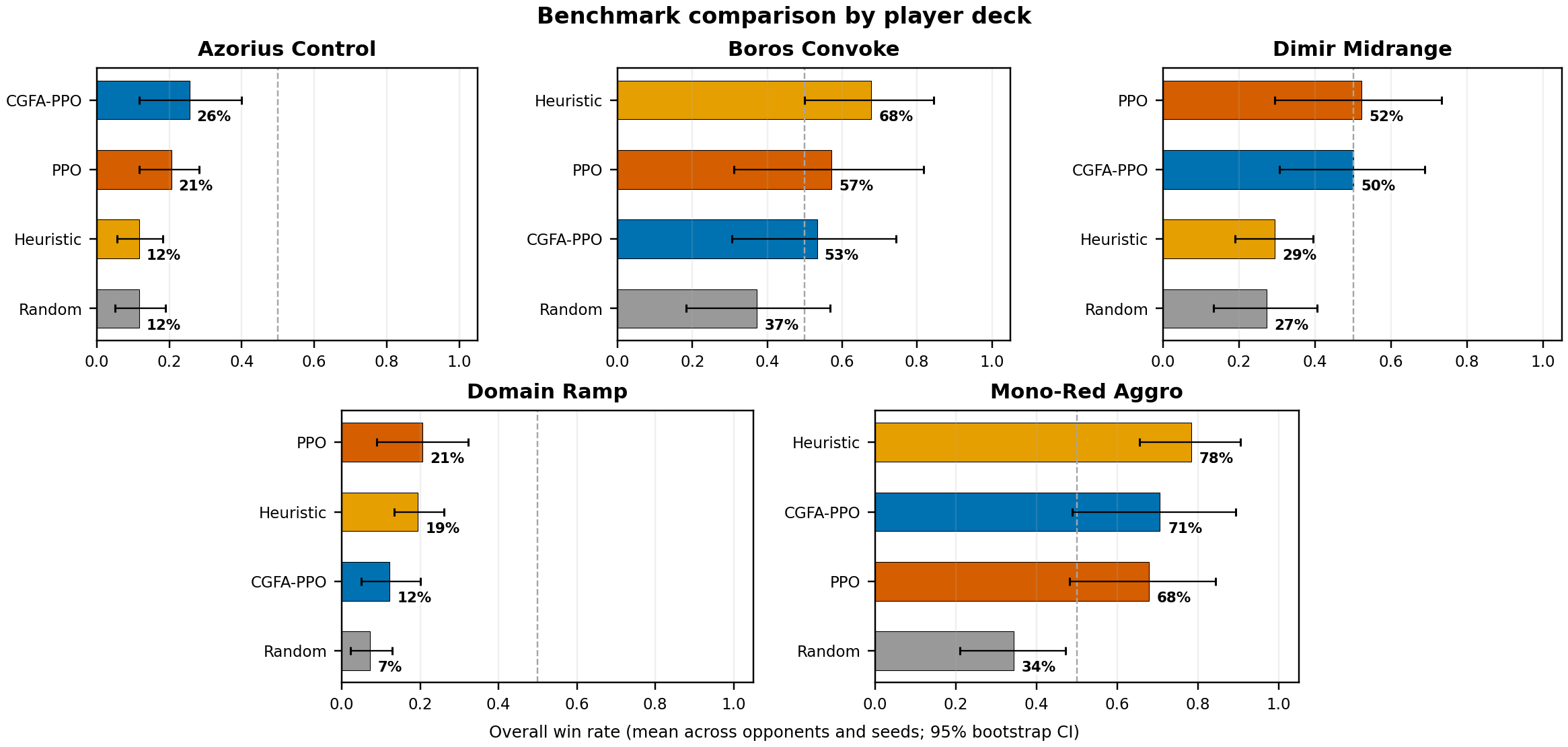}
\caption{Headline benchmark comparison by player deck. Each panel
reports overall win rate against the in-distribution opponent pool
with 95\% percentile-bootstrap CIs over pooled per-seed rates; PPO
and CGFA-PPO are paired across seeds. Random and heuristic policies
are shown as reference anchors for benchmark difficulty.}
\label{fig:headline_winrate}
\end{figure}

\paragraph{Headline outcomes.}
Table~\ref{tab:main_results} reports the per-deck PPO-vs-CGFA-PPO
comparison; Figure~\ref{fig:headline_winrate} adds random and
deck-specialised heuristic anchors. Both learned agents beat random
play across all five decks, but neither dominates uniformly. CGFA-PPO
is higher on Azorius Control ($25.6\%$ vs $20.6\%$) and Mono-Red Aggro
($70.6\%$ vs $67.8\%$); PPO is higher on Boros Convoke, Dimir
Midrange, and Domain Ramp. Heuristics remain strong for proactive
decks but weak for Azorius Control and Domain Ramp, showing that the
benchmark contains both script-favouring and learning-favouring
regimes. At $n_s=2$, Table~\ref{tab:main_results}'s bootstrap columns
are exploratory; their directional message is that CGFA-PPO is
competitive but has not yet produced a uniform headline win-rate gain
over PPO. Additional reference-policy and matchup diagnostics are in
Appendices~\ref{app:additional_results}
and~\ref{app:diagnostics_extra}.

The main positive result is therefore not ``CGFA-PPO wins everywhere'',
but that the benchmark separates several useful behaviours in one
panel. On the aggressive Mono-Red deck, learned policies and the
hand-coded heuristic all perform well, with CGFA-PPO close to the
heuristic anchor and slightly above PPO. On Azorius Control, both
learned agents beat the control heuristic and random policy, but the
absolute win rates remain low, indicating that slower, reactive
archetypes are still difficult under the current training budget. Boros
Convoke and Dimir Midrange sit between these regimes: learned policies
are well above random, yet the heuristic remains competitive on Boros.
Domain Ramp is the clearest failure case for the current learned
agents, where PPO and CGFA-PPO improve over random but remain near the
heuristic and far below the 50\% reference line. These mixed outcomes
are exactly the intended benchmark signal: the suite exposes where
causal structure is competitive, where simple scripts remain strong,
and where longer training or tuning is still needed.

\begin{table}[!htbp]
\centering
\caption{Headline learned-agent results per player deck. Win rates are means over the in-distribution opponent pool and paired seeds; 95\% CIs are percentile-bootstrap intervals over pooled per-opponent seed rates (10k resamples). $n_s$ is the number of paired seeds available; $\Delta$ is the per-deck mean win rate of the test agent minus the reference agent; $p_\text{boot}$ and $p_\text{Holm}$ are the raw and Holm-Bonferroni adjusted paired-bootstrap $p$-values within the headline learned-agent comparison family.}
\label{tab:main_results}
\small
\setlength{\tabcolsep}{4pt}
\begin{tabular}{llcccccc}
\toprule
Deck & Agent & Win Rate (\%) & 95\% CI & $n_s$ & $\Delta$ (pp) & $p_\text{boot}$ & $p_\text{Holm}$ \\
\midrule
Azorius Control & PPO & $20.6$ & $[11.7,\,28.3]$ & $2$ & --- & --- & --- \\
Azorius Control & \textbf{CGFA-PPO} & $\mathbf{25.6}$ & $[11.7,\,40.0]$ & $2$ & $+5.0$ & $0.0001$ & $0.0005$ \\
\midrule
Boros Convoke & PPO & $57.2$ & $[31.1,\,81.7]$ & $2$ & --- & --- & --- \\
Boros Convoke & \textbf{CGFA-PPO} & $\mathbf{53.3}$ & $[29.4,\,74.4]$ & $2$ & $-3.9$ & $0.0001$ & $0.0005$ \\
\midrule
Dimir Midrange & PPO & $52.2$ & $[29.4,\,73.3]$ & $2$ & --- & --- & --- \\
Dimir Midrange & \textbf{CGFA-PPO} & $\mathbf{50.0}$ & $[30.6,\,68.9]$ & $2$ & $-2.2$ & $0.4881$ & $0.9762$ \\
\midrule
Domain Ramp & PPO & $20.6$ & $[9.4,\,32.2]$ & $2$ & --- & --- & --- \\
Domain Ramp & \textbf{CGFA-PPO} & $\mathbf{12.2}$ & $[5.0,\,20.0]$ & $2$ & $-8.3$ & $0.0001$ & $0.0005$ \\
\midrule
Mono-Red Aggro & PPO & $67.8$ & $[48.9,\,84.4]$ & $2$ & --- & --- & --- \\
Mono-Red Aggro & \textbf{CGFA-PPO} & $\mathbf{70.6}$ & $[48.3,\,89.4]$ & $2$ & $+2.8$ & $0.4881$ & $0.9762$ \\
\bottomrule
\end{tabular}
\end{table}

\paragraph{Ablation suite.}
The ablation suite compares masked PPO, CWM-augmented PPO, the
scalar-only control, CGFA-PPO without the gate, CGFA-PPO without
calibration, and full CGFA-PPO. On the Mono-Red diagnostic slice,
scalar-only and CGFA variants cluster near PPO, while the
causal-world-model baseline trails them. This negative evidence shows
that the benchmark separates capacity, factor critics, gating, and
calibration rather than rewarding the label ``causal''.

\begin{figure}[!htbp]
\centering
\includegraphics[width=0.82\linewidth]{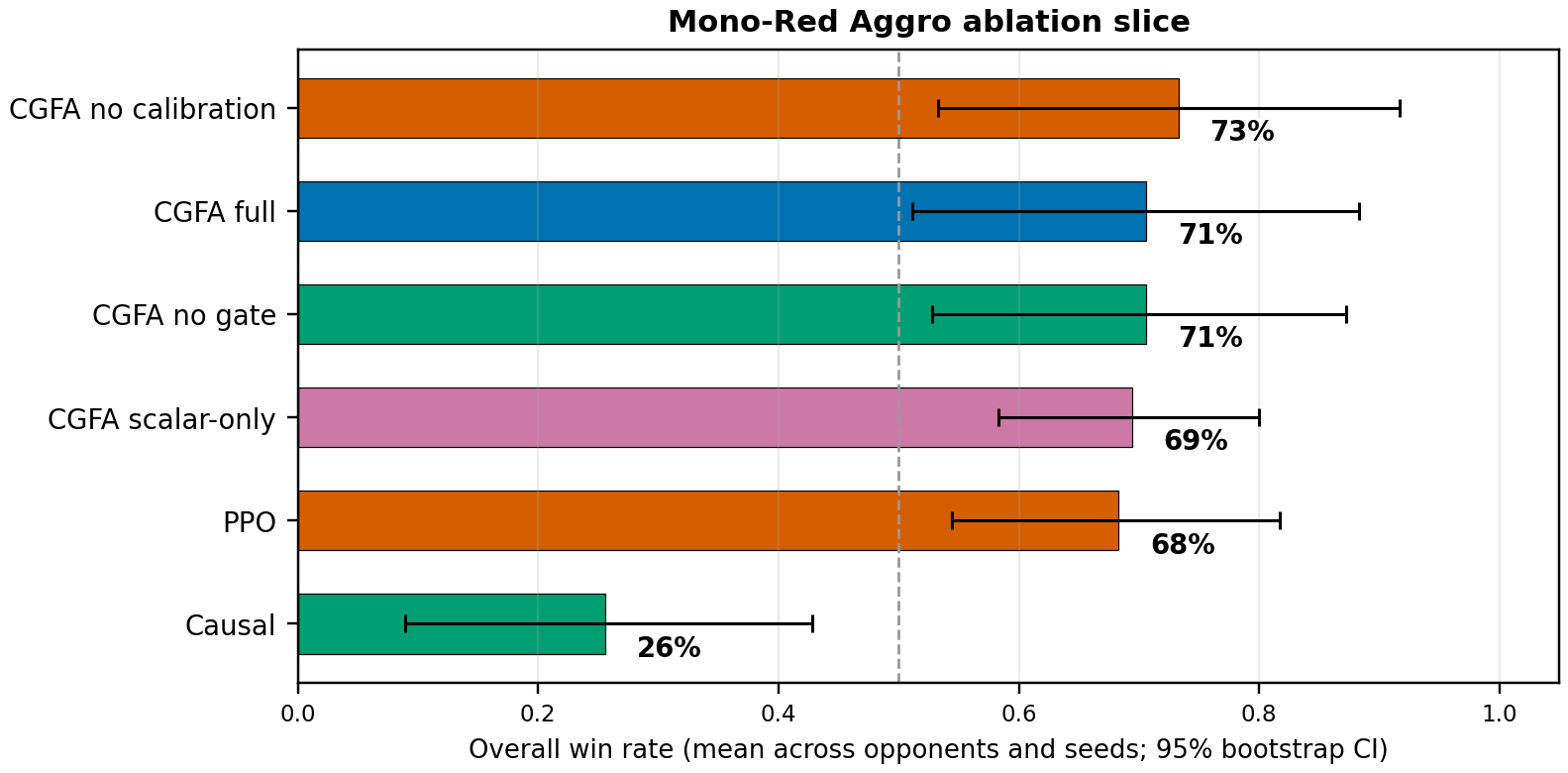}
\caption{Ablation suite on the Mono-Red Aggro diagnostic slice: mean
overall win rate per ablation variant against the in-distribution
opponent pool, with 95\% paired-bootstrap CIs. The variants
disentangle parameter count, factor-aligned critics, the residual
gate, and intervention calibration.}
\label{fig:ablations}
\end{figure}

Table~\ref{tab:ablation} and Figure~\ref{fig:ablations} make the
mechanism picture more concrete. Full CGFA-PPO reaches $70.6\%$, the
no-gate variant is also $70.6\%$, the no-calibration variant reaches
$73.3\%$, and the scalar-only control reaches $69.4\%$, all close to
PPO's $68.3\%$ on this slice. The CWM-augmented causal agent is much
lower at $25.6\%$, despite using causal auxiliary prediction. Thus the
current evidence supports the benchmark interface and the usefulness
of auditing factor heads, but does not show that every causal
augmentation improves scalar win rate. In particular, calibration and
gating are better viewed here as mechanisms for alignment and
interpretability than as already-tuned performance boosters.

\begin{figure}[!htbp]
\centering
\includegraphics[width=0.9\linewidth]{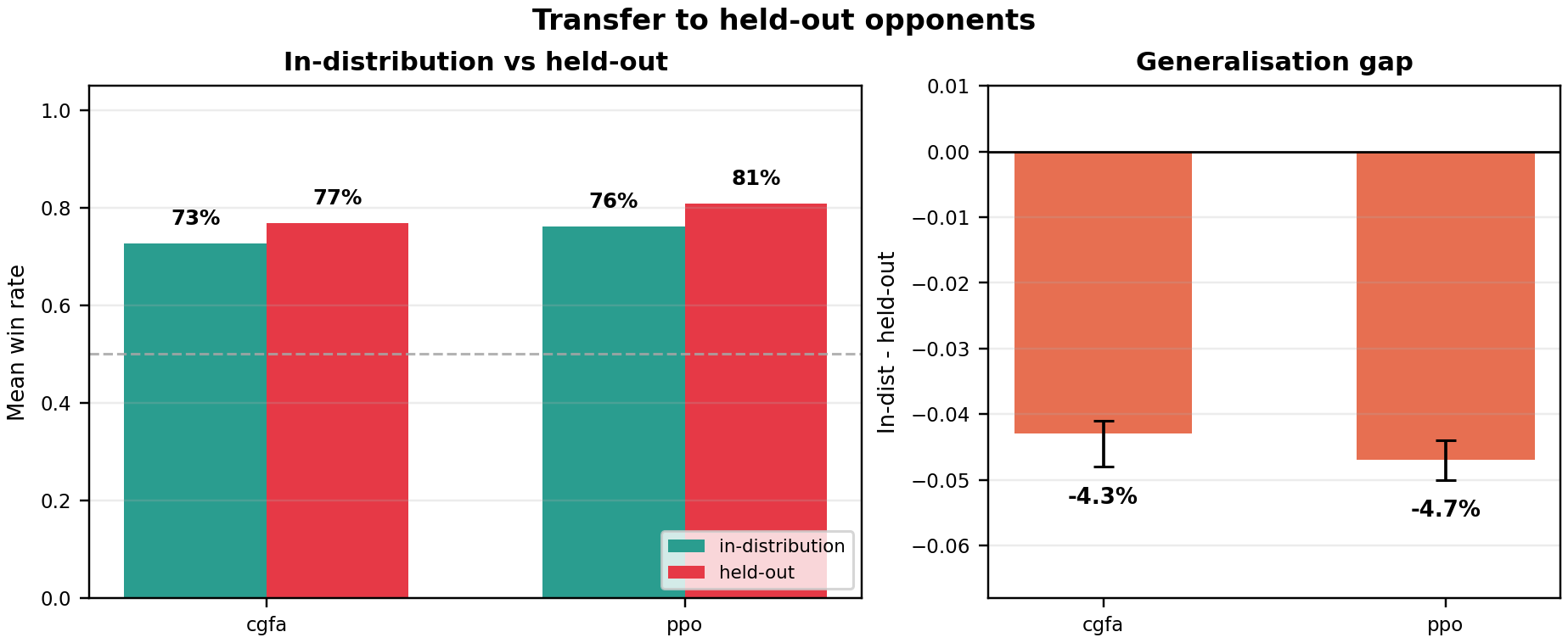}
\caption{Cross-archetype transfer: in-distribution versus held-out
mean win rate per agent, aggregated over paired seed-by-fold
observations. The generalisation gap is the dependent variable for the
per-agent test against zero (Table~\ref{tab:transfer_gap}); negative
values indicate that the held-out opponent mix was easier.}
\label{fig:transfer}
\end{figure}

\paragraph{Cross-archetype transfer.}
The transfer protocol trains on 4 opponents and evaluates on the
held-out archetype. The generalisation gap
$\Delta = \mathrm{WinRate}_{\mathrm{in\text{-}dist}} -
\mathrm{WinRate}_{\mathrm{held\text{-}out}}$ is positive when held-out
performance drops. Table~\ref{tab:transfer_gap} and
Figure~\ref{fig:transfer} show that PPO has a small negative gap
($-4.7$ pp), while CGFA-PPO has a similar negative gap ($-4.3$ pp).
This is a benchmark diagnostic, not evidence that either agent solves
transfer: matchup composition can make held-out opponents softer, and
the low seed count limits inference.

The transfer figure is intentionally read together with the table. PPO
improves from $76.1\%$ in-distribution to $80.8\%$ held-out in this
run, so its negative gap means the held-out opponent mix was easier,
not that transfer is solved. CGFA-PPO shows the same qualitative
pattern, improving from $72.6\%$ to $76.8\%$ with a $[-4.8,-4.1]$ pp
gap interval. This agreement is useful because it demonstrates the
benchmark capability: paired-seed transfer can reveal when held-out
evaluation is easier than the matched training pool and keep that
interpretation separate from scalar win rate.

\begin{table}[!htbp]
\centering
\caption{Cross-archetype generalisation gap per learned agent, paired across seeds. Held-out evaluation aggregates over the leave-one-out folds of the opponent pool; in-distribution evaluation aggregates over the matched training-pool folds. $\Delta = \mathrm{WinRate}_{\mathrm{in\text{-}dist}} - \mathrm{WinRate}_{\mathrm{held\text{-}out}}$ (positive $=$ generalisation drop, negative $=$ held-out is easier). $n_s$ is the number of paired seed-by-fold observations; $p_\text{boot}$ tests $\Delta = 0$ with a paired bootstrap (10k resamples); $p_\text{Holm}$ is the Holm-Bonferroni adjusted $p$-value within the per-agent transfer family.}
\label{tab:transfer_gap}
\small
\setlength{\tabcolsep}{4pt}
\begin{tabular}{lccccccc}
\toprule
Agent & $n_s$ & In-dist (\%) & Held-out (\%) & $\Delta$ (pp) & 95\% CI on $\Delta$ & $p_\text{boot}$ & $p_\text{Holm}$ \\
\midrule
PPO & $2$ & $76.1$ & $80.8$ & $-4.7$ & $[-5.0,\,-4.4]$ & $0.0001$ & $0.0002$ \\
CGFA-PPO & $2$ & $72.6$ & $76.8$ & $-4.3$ & $[-4.8,\,-4.1]$ & $0.0002$ & $0.0003$ \\
\bottomrule
\end{tabular}
\end{table}

\begin{figure}[!htbp]
\centering
\includegraphics[width=\linewidth]{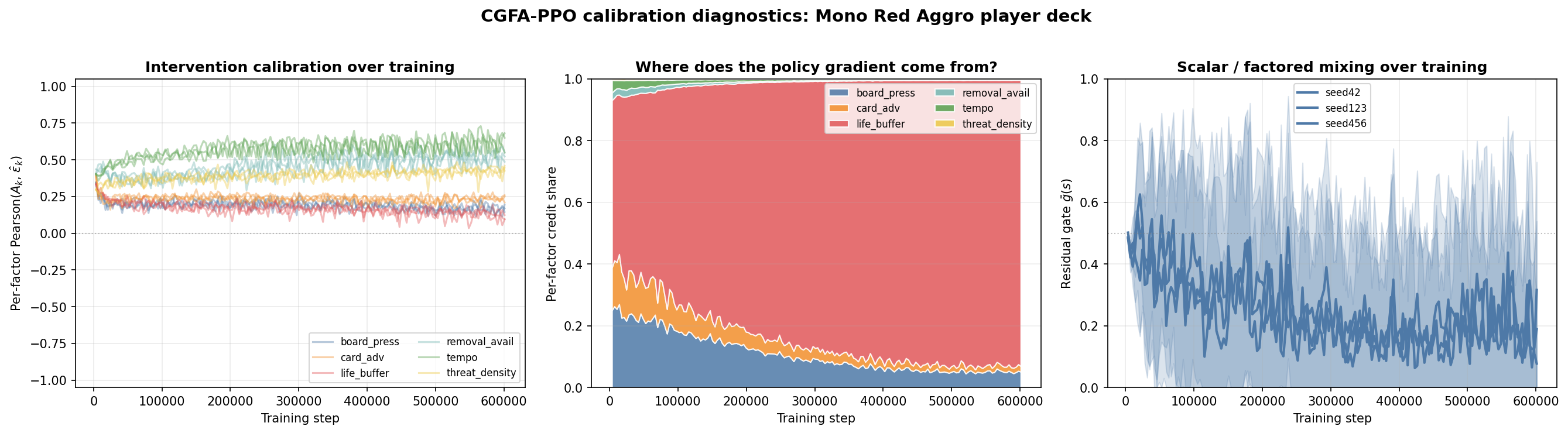}
\caption{CGFA-PPO calibration trajectory for the Mono-Red Aggro
player deck. (a) Per-factor Pearson correlation between $A_k$ and the
SCM-predicted $\varepsilon_k$. (b) Per-factor credit share as a
stacked area. (c) Residual gate $g(s)$ mean (line) with per-state
min/max envelope (shaded).}
\label{fig:calibration}
\end{figure}

\paragraph{Calibration trajectory.}
The benchmark records per-factor calibration, sign agreement, credit
share, and gate statistics. Figure~\ref{fig:calibration} is therefore
a CGFA-PPO mechanism diagnostic, not a PPO-vs-CGFA outcome comparison:
even when scalar win rate is mixed, it shows which SCM factors drive
updates, whether calibration aligns with predicted interventions, and
whether the gate collapses to PPO. A single-episode case study appears
in Appendix~\ref{app:diagnostics_extra}.

In the current calibration diagnostic, several factor correlations move
positive over training, credit share concentrates heavily on the
life-buffer factor, and the residual gate remains active rather than
collapsing entirely to the scalar PPO advantage. These curves are the
main evidence that CGFA-PPO is using the causal interface in an
auditable way even when headline win-rate improvements are
deck-dependent. They also identify concrete follow-up targets:
regularising credit concentration, tuning the gate schedule, and
testing whether stronger calibration improves the difficult control and
ramp archetypes.

\section{Discussion}
\label{sec:discussion}

MTG-Causal-RL measures three benchmark questions in one environment:
\textbf{outcome quality} via win rate and reward, \textbf{generalisation
under opponent shift} via leave-one-out transfer, and \textbf{causal
auditability} via calibration curves and single-episode traces.

\paragraph{Limitations.}
\label{sec:limitations}
The SCM is hand-designed rather than learned, buying auditability at
the cost of structure discovery. The 56-card, 5-archetype pool captures
meaningful diversity at tractable scale, but the full MTG pool is far
larger. Opponents are fixed rather than adaptive, and longer turn caps
increase rollout cost and can shift matchup balance toward slower
archetypes.

\paragraph{Broader impact and language-model agents.}
Card games are recreational; the benchmark poses no direct societal
risk and may support tutorials, strategic-analysis tools, and
auditable decision-making research. The Gymnasium interface also lets
LLM-based strategic agents plug in directly, using SCM factors and
$\mathrm{do}(\cdot)$ queries as compact strategic
summaries~\citep{costarelli2024gamebench, duan2024gtbench}.

\section{Conclusion}
\label{sec:conclusion}

We introduced MTG-Causal-RL, a Gymnasium benchmark on \textit{Magic:
The Gathering} with partial observability, stochastic draws, 478
masked discrete actions, five Standard 2025 archetypes, an explicit
SCM over strategic causal variables, and a paired-seed evaluation
protocol that reports win rate, leave-one-out transfer, calibration,
and causal case studies. We ship a panel of reference baselines and
propose CGFA-PPO as a secondary contribution that exercises the
causal interface. The benchmark establishes MTG as an open, reusable
testbed for causal credit assignment, robust generalisation, and
interpretable sequential decision making.

\bibliographystyle{plainnat}
\bibliography{references}

\newpage
\appendix

\section{Action Space Specification}
\label{app:actions}

Table~\ref{tab:action_space} gives the complete action layout. Total
size: 478 discrete actions. In any given state only 2 to 15 actions
are typically legal; the rest are masked out by the engine.

\begin{table}[ht]
\centering
\caption{Action space layout for MTG-Causal-RL (478 actions across 16
categories). Index ranges are fixed by the benchmark specification.}
\label{tab:action_space}
\small
\begin{tabular}{@{}rrlp{5.4cm}@{}}
\toprule
Index range & Slots & Category & Description \\
\midrule
0        & 1   & PASS                    & End current phase or decline to act \\
1        & 1   & KEEP                    & Keep current opening hand \\
2        & 1   & MULLIGAN                & Shuffle and redraw, drawing one fewer card \\
3        & 1   & CONFIRM                 & Confirm pending action \\
4        & 1   & CANCEL                  & Cancel pending action \\
5        & 1   & AUTO\_PAY               & Automatically pay mana costs \\
6--15    & 10  & BOTTOM                  & Select hand card to bottom (mulligan) \\
16--25   & 10  & DISCARD                 & Discard hand card \\
26--35   & 10  & PLAY\_LAND              & Play hand land \\
36--45   & 10  & CAST\_SORCERY           & Cast sorcery-speed spell from hand \\
46--55   & 10  & CAST\_INSTANT           & Cast instant-speed spell from hand \\
56--115  & 60  & ACTIVATE                & Activate permanent ability \\
116--175 & 60  & ATTACK\_TOGGLE          & Toggle creature as attacker \\
176--235 & 60  & BLOCK\_SELECT\_ATTACKER & Select attacker to assign blocker against \\
236--295 & 60  & BLOCK\_SELECT\_BLOCKER  & Select blocker creature to assign \\
296--417 & 122 & TARGET                  & Select target for spell or ability \\
418--477 & 60  & MANA\_SOURCE            & Select mana source for payment \\
\bottomrule
\end{tabular}
\end{table}

\section{Structural Causal Model: Equations}
\label{app:scm_eqs}

The structural equations underlying the causal graph of
Figure~\ref{fig:scm} are:
\begin{align}
    \mathrm{Mana}_{t+1} &= \mathrm{Mana}_t + \mathbf{1}[\mathrm{LandDrop}_t]
                          + \mathrm{ManaCreatures} \\
    \mathrm{BoardPress} &= \sum_{c \in \mathrm{Own}} \mathrm{power}(c)
                          - \sum_{c \in \mathrm{Opp}} \mathrm{power}(c) \\
    \mathrm{ThreatDensity} &= \frac{|\{c \in \mathrm{Own} :
                                    \mathrm{isThreat}(c)\}|}
                                    {\max(1, |\mathrm{Own}|)} \\
    \mathrm{Tempo} &= \operatorname{clip}\!\left(\frac{\mathrm{ManaSpent}_t}
                                       {\max(1, \mathrm{Mana}_t)} -
                                  \frac{\mathrm{OppManaSpent}_t}
                                       {\max(1, \mathrm{OppMana}_t)},
                                  -1, 1\right) \\
    \mathrm{WinProb} &= \sigma\!\Big(\mathbf{w}^{\!\top}
        [\mathrm{CardAdv}, \mathrm{BoardPress}, \mathrm{Tempo},
         \notag\\&\qquad\;\;
         \mathrm{LifeBuffer}, \mathrm{ThreatDensity},
         \mathrm{RemovalAvail}]\Big),
\end{align}
where $\sigma$ is the sigmoid and $\mathbf{w}$ are learned online from
game outcomes by a logistic-regression head
(Appendix~\ref{app:hyperparams}). The SCM exposes a Pearl-style
$\mathrm{do}(\cdot)$ intervention operator that propagates
interventions through the structural equations using NetworkX-based
descendants enumeration, so the SCM-predicted intervention effect
$\varepsilon_k(s,a) = \phi_k(s') - \phi_k(s)$ used by CGFA-PPO and the
benchmark diagnostics is computed by re-evaluating the equations
after an intervention rather than by simulating the environment.

\section{Reward Shaping with Causal Variables}
\label{app:reward_shaping}

\textit{Potential-based shaping}~\citep{ng1999policy} adds an extra
per-step reward $F(s, s') = \gamma \Phi(s') - \Phi(s)$, where the
potential $\Phi(\cdot)$ is a scalar function of state. In fully
observable MDPs the construction provably preserves the set of
optimal policies for any choice of $\Phi$, so a designer can inject
domain knowledge into the reward signal without distorting what
counts as an optimal policy. The standard policy-invariance proof
assumes that $\Phi$ depends on the full state, which is not available
in a partially observable environment such as MTG. We follow the
extension of~\citet{eck2016potential} and compute $\Phi$ exclusively
from agent-observable quantities, preserving the policy-invariance
argument relative to the agent's own observation process and
preventing the shaped reward from leaking hidden information back
into the policy gradient.

The benchmark's shaped reward uses causal variables as the potential
function:
\begin{align}
    \Phi(s) \;=\; & \alpha_1 \mathrm{Mana}_t + \alpha_2 \mathrm{CardAdv}
                  + \alpha_3 \mathrm{BoardPress} \notag\\
                  & + \alpha_4 \mathrm{Tempo}
                  + \alpha_5 \mathrm{LifeBuffer},\\
    R^{\mathrm{shaped}}_t \;=\; & \gamma\,\Phi(s_{t+1}) - \Phi(s_t)
                  + R^{\mathrm{terminal}}_t.
\end{align}

\section{Hyperparameters}
\label{app:hyperparams}

Table~\ref{tab:hyperparams} lists hyperparameters for all learnable
agents. Values are the benchmark defaults; full reproducibility is
given by the per-run manifest.

\begin{table}[t]
\centering
\caption{Hyperparameters for the PPO, Causal Agent, and CGFA-PPO
baselines.}
\label{tab:hyperparams}
\begin{tabular}{ll}
\toprule
Parameter & Value \\
\midrule
\multicolumn{2}{l}{\textit{Masked PPO backbone (shared)}} \\
Learning rate                  & $3 \times 10^{-4}$, linearly annealed
                                  to $1 \times 10^{-5}$ \\
Minibatch size                 & 256 \\
Rollout steps $n_{\mathrm{steps}}$ & 2{,}048 \\
Epochs per update              & 10 \\
$\gamma$ (discount)            & 0.995 \\
GAE $\lambda$                  & 0.95 \\
Clip range                     & 0.2 \\
Entropy coefficient $c_H$      & 0.05, linearly annealed to 0.005 \\
Max gradient norm              & 0.5 \\
Network architecture           & MLP [512, 256] policy and value
                                 (separate heads), ReLU \\
\midrule
\multicolumn{2}{l}{\textit{Causal Agent (PPO + CWM)}} \\
CWM action embedding dim       & 32 \\
CWM hidden width               & 128 \\
CWM optimiser                  & Adam, learning rate $1 \times 10^{-3}$ \\
CWM replay buffer size         & 20{,}000 transitions \\
CWM training per rollout       & 16 optimiser steps, minibatch 128 \\
CWM aux objective              & $\mathcal{L}_{\mathrm{aux}} +
                                 0.5\,\mathcal{L}_{\mathrm{wp}}$
                                 (MSE on $\Delta\mathrm{CV}$ +
                                 weighted BCE on $g_\phi$) \\
Causal weight $\lambda$        & 0.6 \\
Exploration rate $\epsilon$    & 0.10, annealed to 0.01 \\
\midrule
\multicolumn{2}{l}{\textit{CGFA-PPO (full)}} \\
Number of factors $K$          & 6 (parents of $\mathrm{WinProb}$) \\
Initial blend $\beta$          & SCM logistic-regression weights \\
Per-factor MSE coef $c_f$      & 0.5 \\
Calibration coef $c_c$         & 0.1 \\
Residual gate                  & state-conditional MLP, hidden 32 \\
Initial gate value $g_0$       & 0.5 \\
Gate entropy coef $c_e$        & 0.0 (default; sweep value
                                  in ablation) \\
\midrule
\multicolumn{2}{l}{\textit{WinProbLearner (SCM)}} \\
Outcome buffer                 & 2{,}000 terminal games \\
Refit interval                 & 200 games (200 SGD steps,
                                  $\eta = 0.01$) \\
Feature standardisation        & yes (z-scored prior to fit) \\
\midrule
\multicolumn{2}{l}{\textit{Heuristic agents (per archetype)}} \\
Mulligan land range            & $[1,5]$ for aggro,
                                 $[2,5]$ for control / midrange / ramp \\
Aggression                     & varies by archetype \\
Defensive life threshold       & 8 (control), N/A (aggro) \\
\bottomrule
\end{tabular}
\end{table}

\section{CGFA-PPO Algorithm Specification}
\label{app:cgfa_algorithm}

This appendix gives the formal specification of the CGFA-PPO reference
agent introduced in Section~\ref{sec:cgfa}. The main paper text gives
the high-level architecture and the role of each component; here we
list the per-step quantities published by the benchmark wrapper, the
per-factor critic and GAE construction, the residual blending rule,
the intervention-calibration loss, and the full training objective.

\subsection{Architecture and per-step quantities}
\label{app:cgfa_arch}

\begin{figure}[t]
\centering
\resizebox{\textwidth}{!}{%
\begin{tikzpicture}[
    node distance=10mm,
    every node/.style={font=\small},
    box/.style={draw, rounded corners, fill=blue!8, align=center,
                inner sep=3pt, minimum height=8mm},
    head/.style={draw, rounded corners=1mm, fill=green!15, align=center,
                 inner sep=3pt, minimum height=6mm},
    gate/.style={draw, rounded corners=1mm, fill=orange!20, align=center,
                 inner sep=3pt, minimum height=6mm},
    cal/.style={draw, dashed, rounded corners=1mm, fill=red!10, align=center,
                inner sep=3pt, minimum height=6mm, font=\footnotesize},
    arr/.style={-{Stealth[length=1.5mm]}, thin},
]
    \node[box] (obs) {$s_t \in \mathbb{R}^{3077}$};
    \node[box, right=of obs] (mlp) {Critic\\MLP};
    \node[box, right=of mlp] (latent) {$h(s_t)$};

    \node[head, above right=2mm and 12mm of latent] (vscalar) {$V(s_t)$\\(scalar)};
    \node[head, right=of latent] (vfactors) {$V_k(s_t)$\\for $k=1\ldots K$};
    \node[gate, below right=2mm and 12mm of latent] (gnode) {$g(s_t) \in (0,1)$};

    \node[box, right=14mm of vfactors,
          fill=green!8] (advs) {$A_{\mathrm{used}} = (1-g)\,A^{\mathrm{scalar}}
                                + g\sum_k w_k A_k$};

    \node[cal, below=10mm of advs] (cal) {Pearson calibration loss\\
                                          \scriptsize{maximise
                                          $\mathrm{corr}(A_k,
                                          \varepsilon_k)$ per factor}};

    \node[box, right=of advs, fill=blue!12] (ppo) {Masked PPO\\surrogate};

    \draw[arr] (obs)     -- (mlp);
    \draw[arr] (mlp)     -- (latent);
    \draw[arr] (latent)  -- (vscalar);
    \draw[arr] (latent)  -- (vfactors);
    \draw[arr] (latent)  -- (gnode);
    \draw[arr] (vscalar) -- (advs.north west);
    \draw[arr] (vfactors) -- (advs);
    \draw[arr] (gnode)   -- (advs.south west);
    \draw[arr] (advs)    -- (ppo);
    \draw[arr, dashed]
        (vfactors.south) to[bend right=15] (cal.north);
    \draw[arr, dashed] (cal) -- (ppo.south west);
\end{tikzpicture}%
}
\caption{Reference CGFA-PPO agent used in MTG-Causal-RL. The benchmark
supplies causal factors and SCM-predicted intervention effects. The
agent preserves the scalar PPO critic, adds per-factor value heads,
and uses a state-conditional gate to blend scalar and factored
advantages. The calibration loss measures whether the learned factor
advantages align with the SCM intervention target.}
\label{fig:cgfa}
\end{figure}
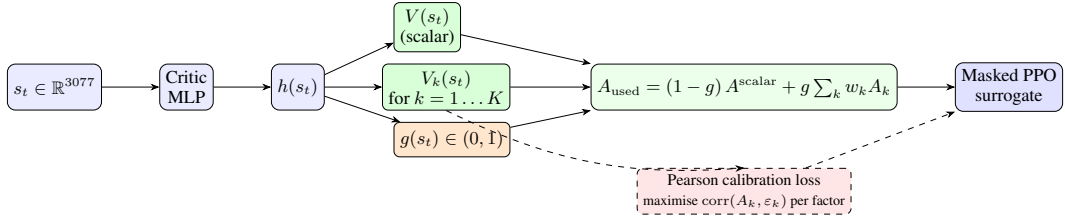

CGFA-PPO is a drop-in replacement for publication
PPO~\citep{schulman2017proximal, huang2022masking} that augments the
critic without removing it. Let $\phi(s) \in \mathbb{R}^{K}$ denote the
$K$ benchmark causal factors, defined as the parents of
$\mathrm{WinProb}$ in Table~\ref{tab:causal_vars}; for the default
benchmark configuration $K = 6$. At each environment step, the CGFA
Gymnasium wrapper publishes
\begin{itemize}
    \item $\phi(s_t) \in \mathbb{R}^{K}$, the per-factor values;
    \item $r^{\mathrm{factor}}_{k,t} = \phi_k(s_{t+1}) - \phi_k(s_t)$,
    the per-factor reward used as the GAE target for head $k$;
    \item $\varepsilon_{k,t} =
    \phi^{\mathrm{SCM}}_{k}(s_{t+1}) - \phi^{\mathrm{SCM}}_{k}(s_t)$,
    the SCM-predicted per-factor change after re-evaluating the
    structural equations of Appendix~\ref{app:scm_eqs}.
\end{itemize}
The policy network has the standard masked PPO actor and a scalar
critic $V(s)$, plus a causal value head that emits
$V_k(s) \in \mathbb{R}^{K}$ from the same critic features and a
state-conditional residual gate $g(s) \in (0, 1)$. The mixture weights
$w_k = \mathrm{softmax}(\beta)_k$ over the $K$ factor heads are
themselves learnable parameters, initialised from the WinProbLearner's
logistic-regression weights so that CGFA-PPO starts pre-loaded with the
structural prior. Figure~\ref{fig:cgfa} sketches the architecture.

\subsection{Per-factor critic and GAE}
\label{app:cgfa_critic}

For factor $k$, the per-factor return target is
\begin{equation}
    G_{k,t} \;=\; \sum_{i \geq 0} \gamma^{i}\, r^{\mathrm{factor}}_{k,t+i},
\end{equation}
with the same discount factor $\gamma$ as the scalar return. Training
applies a standard mean-squared loss
\begin{equation}
    \mathcal{L}_{\mathrm{factor}} \;=\;
    \frac{1}{K} \sum_{k=1}^{K}
    \mathbb{E}_t \big[\big(G_{k,t} - V_k(s_t)\big)^{2}\big]
\end{equation}
on top of the standard scalar value loss, weighted by $\lambda_{f}$.
The per-factor advantage
\begin{equation}
    A_{k,t} \;=\; G_{k,t} - V_k(s_t)
\end{equation}
is computed using the same generalised-advantage truncation as the
scalar critic in standard GAE~\citep{schulman2016gae}, which itself
rests on the temporal-difference targets of~\citet{sutton1988learning}.
We use $A_k$ only as an input to the residual blend
(Section~\ref{app:cgfa_blend}). Per-factor returns and advantages are
also stored in the rollout buffer so that the
intervention-calibration loss has access to them.

\subsection{State-conditional residual gate}
\label{app:cgfa_blend}

The blended advantage actually fed to the PPO surrogate is
\begin{equation}
    A_{\mathrm{used}}(s_t, a_t) \;=\;
    \big(1 - g(s_t)\big)\, A^{\mathrm{scalar}}(s_t, a_t)
    \;+\; g(s_t) \sum_{k=1}^{K} w_k\, A_k(s_t, a_t),
    \label{eq:cgfa_blend_appendix}
\end{equation}
where $g(s_t) \in (0, 1)$ is produced by a small MLP over the critic
latent features and $w_k = \mathrm{softmax}(\beta)_k$ are the
learnable mixture weights. When $g(s_t) \to 0$, CGFA-PPO reduces to
vanilla masked PPO; when $g(s_t) \to 1$, the policy update relies
entirely on the structured factor-aligned advantage. Letting $g$
depend on $s_t$ allows the policy to decide, per state, whether the
SCM-aligned signal is informative enough to trust at that decision
point. The benchmark logs the per-state $g(s_t)$ distribution and the
per-factor credit share
$|w_k A_k| / \sum_{j} |w_j A_j|$ per update.

\subsection{Intervention-calibration loss}
\label{app:cgfa_calibration}

The structural prior provided by the SCM is encoded as a
Pearson-correlation auxiliary loss between each factor's per-state
advantage and the SCM's predicted intervention effect:
\begin{equation}
    \mathcal{L}_{\mathrm{cal}} \;=\;
    -\frac{1}{K} \sum_{k=1}^{K}
    \frac{\operatorname{Cov}_t\!\big(A_{k,t}, \varepsilon_{k,t}\big)}
         {\sigma_{A_k}\, \sigma_{\varepsilon_k} + \delta},
\end{equation}
with $\delta$ a small constant for numerical stability and a clamp on
the per-factor standard deviation to avoid amplification when a factor
has near-zero variance in a minibatch. The diagnostic value of this
loss is as important as its training signal: the benchmark reports
whether each factor advantage moves in the same direction as the SCM's
intervention target, whether the gate collapses to scalar PPO, and
which causal factors dominate the policy update over training.

\subsection{Full training objective}
\label{app:cgfa_objective}

Let $\mathcal{L}_{\mathrm{ppo}}$ denote the standard masked PPO clipped
surrogate computed on $A_{\mathrm{used}}$,
$\mathcal{L}_{\mathrm{value}}$ the scalar critic loss,
$\mathcal{L}_{\mathrm{ent}}$ the masked-policy entropy bonus, and
$\mathcal{L}_{\mathrm{gate}}$ a small per-state entropy bonus on
$g(s_t)$ that prevents premature collapse to either extreme. The full
CGFA-PPO objective minimised at each update is
\begin{equation}
    \mathcal{L}_{\mathrm{CGFA}} \;=\;
    \mathcal{L}_{\mathrm{ppo}}
    \;+\; c_{v}\, \mathcal{L}_{\mathrm{value}}
    \;-\; c_{H}\, \mathcal{L}_{\mathrm{ent}}
    \;+\; c_{f}\, \mathcal{L}_{\mathrm{factor}}
    \;+\; c_{c}\, \mathcal{L}_{\mathrm{cal}}
    \;-\; c_{e}\, \mathcal{L}_{\mathrm{gate}},
\end{equation}
with coefficients $c_v$, $c_H$, $c_f$, $c_c$, $c_e$ listed in
Table~\ref{tab:hyperparams}. The ablation suite of
Section~\ref{sec:experiments} disables $\mathcal{L}_{\mathrm{factor}}$,
$\mathcal{L}_{\mathrm{cal}}$, and $g(s_t)$ in turn to isolate the
contribution of each component.

\subsection{Training-loop pseudocode}
\label{app:cgfa_pseudocode}

Algorithm~\ref{alg:cgfa_ppo} ties the per-step quantities of
Section~\ref{app:cgfa_arch}, the per-factor critic of
Section~\ref{app:cgfa_critic}, the residual blend of
Section~\ref{app:cgfa_blend}, the calibration loss of
Section~\ref{app:cgfa_calibration}, and the full objective of
Section~\ref{app:cgfa_objective} into a single one-update training
loop. Quantities subscripted by $k$ run over the $K$ benchmark causal
factors; quantities subscripted by $t$ run over rollout timesteps.

\begin{algorithm}[t]
\caption{CGFA-PPO: one update.}
\label{alg:cgfa_ppo}
\begin{algorithmic}[1]
\Require Policy $\pi_\theta$ (masked PPO actor); scalar critic
$V_\theta(s)$; per-factor critic $V_{k,\theta}(s)$ for $k=1,\dots,K$;
gate $g_\theta(s) \in (0,1)$; mixture logits
$\beta \in \mathbb{R}^{K}$; coefficients $c_v, c_H, c_f, c_c, c_e$;
discount $\gamma$; GAE parameter $\lambda$; rollout length $T$.
\State Reset rollout buffer $\mathcal{B} \gets \varnothing$; sample
initial state $s_0$ from the environment.
\For{$t = 0, \dots, T-1$} \Comment{collect rollout}
  \State Read action mask $m_t$ from the environment.
  \State $a_t \sim \pi_\theta(\cdot \mid s_t, m_t)$; observe
  $r_t, s_{t+1}$.
  \State Read per-factor values $\phi(s_t), \phi(s_{t+1})$ and
  per-factor SCM-predicted change $\varepsilon_t \in \mathbb{R}^{K}$
  from the CGFA wrapper.
  \State Per-factor reward $r^{\mathrm{factor}}_{k,t} \gets \phi_k(s_{t+1}) - \phi_k(s_t)$ for $k=1,\dots,K$.
  \State Store
  $\big(s_t, m_t, a_t, r_t, r^{\mathrm{factor}}_{k,t}, \varepsilon_{k,t},
  V_\theta(s_t), V_{k,\theta}(s_t), g_\theta(s_t)\big)$ in $\mathcal{B}$.
\EndFor
\State Compute scalar GAE
$A^{\mathrm{scalar}}_t$ from $r_t$ and $V_\theta(s_t)$ using
$(\gamma, \lambda)$.
\State Compute per-factor returns
$G_{k,t} \gets \sum_{i \geq 0} \gamma^{i}\, r^{\mathrm{factor}}_{k,t+i}$ and
per-factor advantages $A_{k,t} \gets G_{k,t} - V_{k,\theta}(s_t)$
under the same GAE truncation.
\State Mixture weights $w_k \gets \mathrm{softmax}(\beta)_k$;
blended advantage $A^{\mathrm{used}}_t \gets
\big(1 - g_\theta(s_t)\big)\, A^{\mathrm{scalar}}_t
+ g_\theta(s_t) \sum_{k=1}^{K} w_k\, A_{k,t}$.
\State Normalise $A^{\mathrm{used}}_t$ across the rollout (zero mean,
unit variance) before PPO updates.
\For{epoch $= 1, \dots, E$}
  \For{minibatch $\mathcal{M} \subset \mathcal{B}$}
    \State $\mathcal{L}_{\mathrm{ppo}} \gets$ masked PPO clipped
    surrogate on $A^{\mathrm{used}}_t$ over $\mathcal{M}$.
    \State $\mathcal{L}_{\mathrm{value}} \gets
    \mathbb{E}_{\mathcal{M}}\big[(R_t - V_\theta(s_t))^{2}\big]$ with
    bootstrapped return $R_t$.
    \State $\mathcal{L}_{\mathrm{factor}} \gets
    \tfrac{1}{K}\sum_{k=1}^{K}
    \mathbb{E}_{\mathcal{M}}\big[(G_{k,t} - V_{k,\theta}(s_t))^{2}\big]$.
    \State $\mathcal{L}_{\mathrm{cal}} \gets
    -\tfrac{1}{K}\sum_{k=1}^{K}
    \dfrac{\operatorname{Cov}_{\mathcal{M}}(A_{k,t}, \varepsilon_{k,t})}
          {\sigma_{A_k}\, \sigma_{\varepsilon_k} + \delta}$.
    \State $\mathcal{L}_{\mathrm{ent}} \gets$ mask-aware policy
    entropy on $\mathcal{M}$;
    $\mathcal{L}_{\mathrm{gate}} \gets$ Bernoulli entropy of
    $g_\theta(s_t)$ on $\mathcal{M}$.
    \State Total loss
    $\mathcal{L}_{\mathrm{CGFA}} \gets
    \mathcal{L}_{\mathrm{ppo}}
    + c_v\, \mathcal{L}_{\mathrm{value}}
    - c_H\, \mathcal{L}_{\mathrm{ent}}
    + c_f\, \mathcal{L}_{\mathrm{factor}}
    + c_c\, \mathcal{L}_{\mathrm{cal}}
    - c_e\, \mathcal{L}_{\mathrm{gate}}$.
    \State $\theta, \beta \gets \mathrm{Adam}\big(\theta, \beta;\,
    \nabla \mathcal{L}_{\mathrm{CGFA}}\big)$ with global gradient norm
    clipped to $g_{\max}$.
  \EndFor
\EndFor
\State Log per-factor calibration metrics: per-factor Pearson
correlation $\operatorname{corr}(A_{k,t}, \varepsilon_{k,t})$,
per-factor credit share $|w_k A_k| / \sum_{j} |w_j A_j|$, and
per-state gate distribution of $g_\theta(s_t)$.
\end{algorithmic}
\end{algorithm}

\section{Statistical Methodology}
\label{app:stats}

The benchmark aggregation protocol provides the statistical primitives
used throughout Section~\ref{sec:experiments}.

\paragraph{Win rate confidence intervals.}
Per-cell win rates use Wilson score
intervals~\citep{wilson1927probable}, which give correct coverage at
small $n$ even when the empirical proportion is close to 0 or 1.
Mean win rates across seeds use BCa paired
bootstrap~\citep{efron1993bootstrap} with $B = 10{,}000$ resamples.

\paragraph{Pairwise tests.}
Pairwise comparisons across agents use Welch
$t$-tests~\citep{welch1947generalization} (CIs computed from
Welch-Satterthwaite degrees of freedom) and Wilcoxon signed-rank
tests~\citep{wilcoxon1945individual}; the Wilcoxon CI is computed from
a bootstrap on the Hodges-Lehmann location
estimator~\citep{hodges1963lehmann} of paired
differences. Multiple-comparison correction across the ablation suite
is Holm-Bonferroni~\citep{holm1979simple}.

\paragraph{Causal-agent calibration metrics.}
The Pearson correlation~\citep{pearson1895correlation} between $A_k$
and $\varepsilon_k$ is reported per factor per update, together with
the sign-agreement rate
$\mathrm{Pr}(\mathrm{sign}(A_k) = \mathrm{sign}(\varepsilon_k))$
restricted to $\{(A_k, \varepsilon_k) : A_k \neq 0,
\varepsilon_k \neq 0\}$, the per-state gate distribution, and the
per-factor credit share $|w_k A_k| / \sum_j |w_j A_j|$. These metrics
power Figure~\ref{fig:calibration}.

\section{Additional Benchmark Tables}
\label{app:additional_results}

The main-text headline comparison
(Table~\ref{tab:main_results}, Figure~\ref{fig:headline_winrate})
isolates the learned-agent contrast. This appendix completes the
picture with fixed reference policies and the fully expanded
per-deck and ablation rows. Table~\ref{tab:reference_anchors}
reports the random and deck-specialised heuristic policies used as
reference anchors. Table~\ref{tab:headline} reports the per-deck
learned-agent rows underlying the main-text aggregates, and
Table~\ref{tab:ablation} reports the detailed ablation and
reference-policy rows underlying Figure~\ref{fig:ablations}.

\begin{table}[t]
\centering
\caption{Reference-policy anchors for interpreting headline win rates. These
policies are fixed rather than trained, so they calibrate benchmark difficulty
instead of serving as the primary learned-agent comparison.}
\label{tab:reference_anchors}
\begin{tabular}{llcc}
\toprule
Policy & Player Deck & Win Rate & 95\% CI \\
\midrule
Heuristic & Azorius Control & $0.114$ & $[0.071,\,0.157]$ \\
Heuristic & Mono-Red Aggro & $0.790$ & $[0.672,\,0.903]$ \\
Random & Azorius Control & $0.123$ & $[0.064,\,0.189]$ \\
Random & Mono-Red Aggro & $0.436$ & $[0.291,\,0.589]$ \\
\bottomrule
\end{tabular}
\end{table}

\begin{table}[t]
\centering
\caption{Overall win rate (mean and 95\% percentile bootstrap CI on pooled seed-level rates, 10k resamples).}
\label{tab:headline}
\begin{tabular}{llll}
\toprule
Source & Agent & Player Deck & Win Rate (95\% CI) \\
\midrule
Ablation: Causal & Causal & Mono-Red Aggro & $0.256\;[0.089,\,0.428]$ \\
Ablation: Causal & Heuristic & Mono-Red Aggro & $0.783\;[0.656,\,0.906]$ \\
Ablation: Causal & Random & Mono-Red Aggro & $0.350\;[0.217,\,0.483]$ \\
Ablation: CGFA-PPO & CGFA-PPO & Mono-Red Aggro & $0.706\;[0.511,\,0.883]$ \\
Ablation: CGFA-PPO & Heuristic & Mono-Red Aggro & $0.783\;[0.656,\,0.906]$ \\
Ablation: CGFA-PPO & Random & Mono-Red Aggro & $0.344\;[0.211,\,0.472]$ \\
Ablation: no calibration & CGFA-PPO & Mono-Red Aggro & $0.733\;[0.533,\,0.917]$ \\
Ablation: no calibration & Heuristic & Mono-Red Aggro & $0.783\;[0.656,\,0.906]$ \\
Ablation: no calibration & Random & Mono-Red Aggro & $0.344\;[0.211,\,0.472]$ \\
Ablation: no gate & CGFA-PPO & Mono-Red Aggro & $0.706\;[0.528,\,0.872]$ \\
Ablation: no gate & Heuristic & Mono-Red Aggro & $0.783\;[0.656,\,0.906]$ \\
Ablation: no gate & Random & Mono-Red Aggro & $0.356\;[0.222,\,0.483]$ \\
Ablation: scalar-only & CGFA scalar-only & Mono-Red Aggro & $0.694\;[0.583,\,0.800]$ \\
Ablation: scalar-only & Heuristic & Mono-Red Aggro & $0.783\;[0.656,\,0.906]$ \\
Ablation: scalar-only & Random & Mono-Red Aggro & $0.344\;[0.211,\,0.472]$ \\
Ablation: PPO & Heuristic & Mono-Red Aggro & $0.783\;[0.656,\,0.906]$ \\
Ablation: PPO & PPO & Mono-Red Aggro & $0.683\;[0.544,\,0.817]$ \\
Ablation: PPO & Random & Mono-Red Aggro & $0.367\;[0.217,\,0.522]$ \\
Headline & CGFA-PPO & Azorius Control & $0.256\;[0.117,\,0.400]$ \\
Headline & CGFA-PPO & Boros Convoke & $0.533\;[0.306,\,0.744]$ \\
Headline & CGFA-PPO & Dimir Midrange & $0.500\;[0.306,\,0.689]$ \\
Headline & CGFA-PPO & Domain Ramp & $0.122\;[0.050,\,0.200]$ \\
Headline & CGFA-PPO & Mono-Red Aggro & $0.706\;[0.489,\,0.894]$ \\
Headline & Heuristic & Azorius Control & $0.117\;[0.056,\,0.183]$ \\
Headline & Heuristic & Boros Convoke & $0.678\;[0.500,\,0.844]$ \\
Headline & Heuristic & Mono-Red Aggro & $0.783\;[0.656,\,0.906]$ \\
Headline & Heuristic & Dimir Midrange & $0.294\;[0.189,\,0.394]$ \\
Headline & PPO & Azorius Control & $0.206\;[0.117,\,0.289]$ \\
Headline & PPO & Boros Convoke & $0.572\;[0.311,\,0.822]$ \\
Headline & PPO & Dimir Midrange & $0.522\;[0.294,\,0.733]$ \\
Headline & PPO & Domain Ramp & $0.206\;[0.089,\,0.322]$ \\
Headline & PPO & Mono-Red Aggro & $0.678\;[0.483,\,0.844]$ \\
Headline & Heuristic & Domain Ramp & $0.194\;[0.133,\,0.261]$ \\
Headline & Random & Azorius Control & $0.117\;[0.050,\,0.189]$ \\
Headline & Random & Boros Convoke & $0.372\;[0.183,\,0.567]$ \\
Headline & Random & Dimir Midrange & $0.272\;[0.133,\,0.406]$ \\
Headline & Random & Domain Ramp & $0.072\;[0.022,\,0.128]$ \\
Headline & Random & Mono-Red Aggro & $0.344\;[0.211,\,0.472]$ \\
Transfer held-out & CGFA-PPO & Mono-Red Aggro & $0.758\;[0.650,\,0.867]$ \\
Transfer held-out & PPO & Mono-Red Aggro & $0.808\;[0.717,\,0.900]$ \\
Transfer in-dist & CGFA-PPO & Mono-Red Aggro & $0.783\;[0.672,\,0.889]$ \\
Transfer in-dist & PPO & Mono-Red Aggro & $0.761\;[0.650,\,0.867]$ \\
\bottomrule
\end{tabular}
\end{table}

\begin{table}[t]
\centering
\caption{Overall win rate (mean and 95\% percentile bootstrap CI on pooled seed-level rates, 10k resamples).}
\label{tab:ablation}
\begin{tabular}{llll}
\toprule
Source & Agent & Player Deck & Win Rate (95\% CI) \\
\midrule
Causal & Causal & Mono-Red Aggro & $0.256\;[0.089,\,0.428]$ \\
Causal & Heuristic & Mono-Red Aggro & $0.783\;[0.656,\,0.906]$ \\
Causal & Random & Mono-Red Aggro & $0.350\;[0.217,\,0.483]$ \\
CGFA-PPO & CGFA-PPO & Mono-Red Aggro & $0.706\;[0.511,\,0.883]$ \\
CGFA-PPO & Heuristic & Mono-Red Aggro & $0.783\;[0.656,\,0.906]$ \\
CGFA-PPO & Random & Mono-Red Aggro & $0.344\;[0.211,\,0.472]$ \\
CGFA no calibration & CGFA-PPO & Mono-Red Aggro & $0.733\;[0.533,\,0.917]$ \\
CGFA no calibration & Heuristic & Mono-Red Aggro & $0.783\;[0.656,\,0.906]$ \\
CGFA no calibration & Random & Mono-Red Aggro & $0.344\;[0.211,\,0.472]$ \\
CGFA no gate & CGFA-PPO & Mono-Red Aggro & $0.706\;[0.528,\,0.872]$ \\
CGFA no gate & Heuristic & Mono-Red Aggro & $0.783\;[0.656,\,0.906]$ \\
CGFA no gate & Random & Mono-Red Aggro & $0.356\;[0.222,\,0.483]$ \\
CGFA scalar-only & CGFA scalar-only & Mono-Red Aggro & $0.694\;[0.583,\,0.800]$ \\
CGFA scalar-only & Heuristic & Mono-Red Aggro & $0.783\;[0.656,\,0.906]$ \\
CGFA scalar-only & Random & Mono-Red Aggro & $0.344\;[0.211,\,0.472]$ \\
PPO & Heuristic & Mono-Red Aggro & $0.783\;[0.656,\,0.906]$ \\
PPO & PPO & Mono-Red Aggro & $0.683\;[0.544,\,0.817]$ \\
PPO & Random & Mono-Red Aggro & $0.367\;[0.217,\,0.522]$ \\
\bottomrule
\end{tabular}
\end{table}

\section{Additional Diagnostic Figures}
\label{app:diagnostics_extra}

\begin{figure}[!htbp]
\centering
\includegraphics[width=\linewidth]{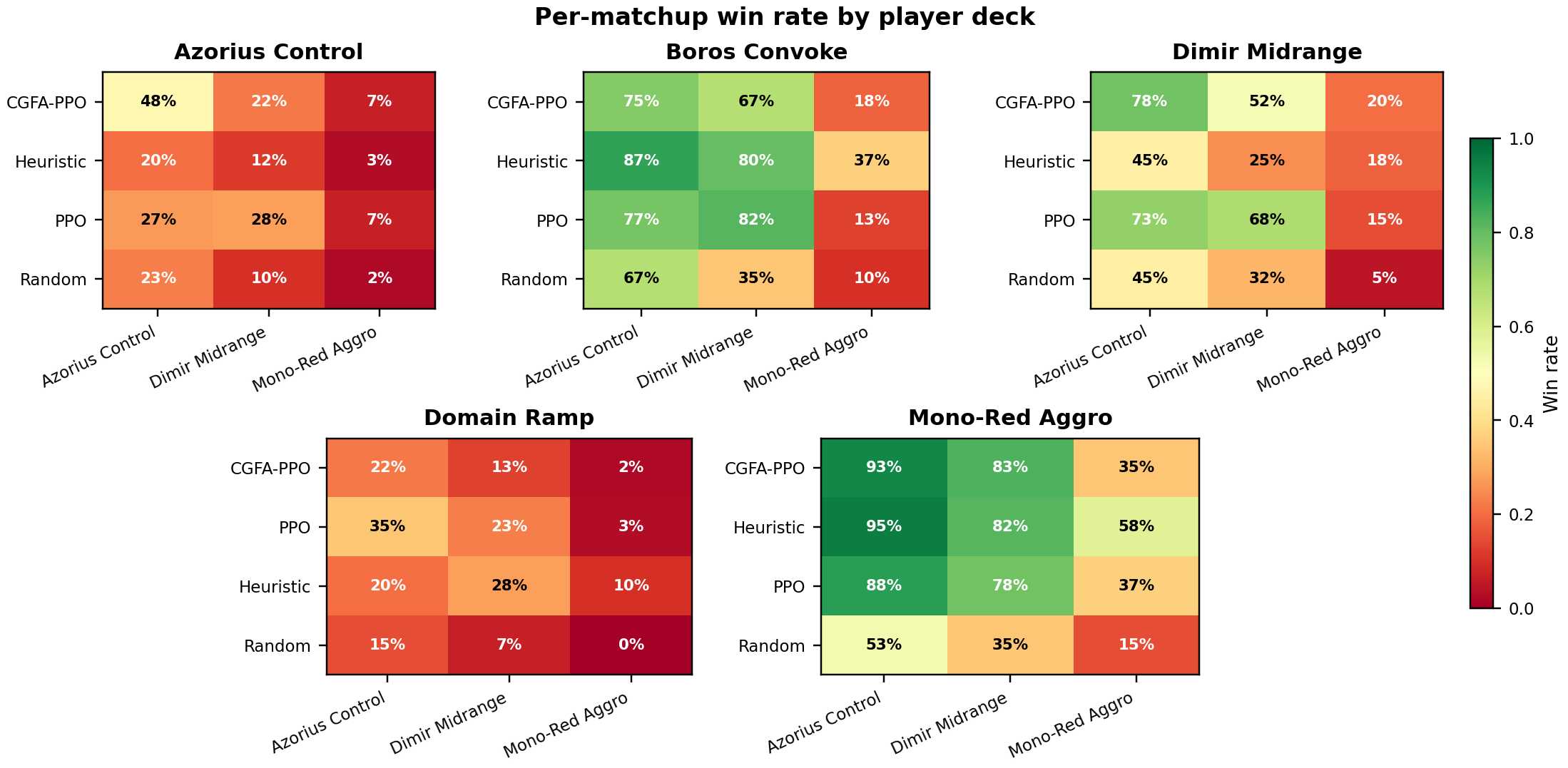}
\caption{Headline benchmark diagnostics: per-matchup win-rate heatmap
across agents, player decks, and in-distribution opponents.}
\label{fig:headline_diagnostics}
\end{figure}

\begin{figure}[!htbp]
\centering
\includegraphics[width=0.95\linewidth]{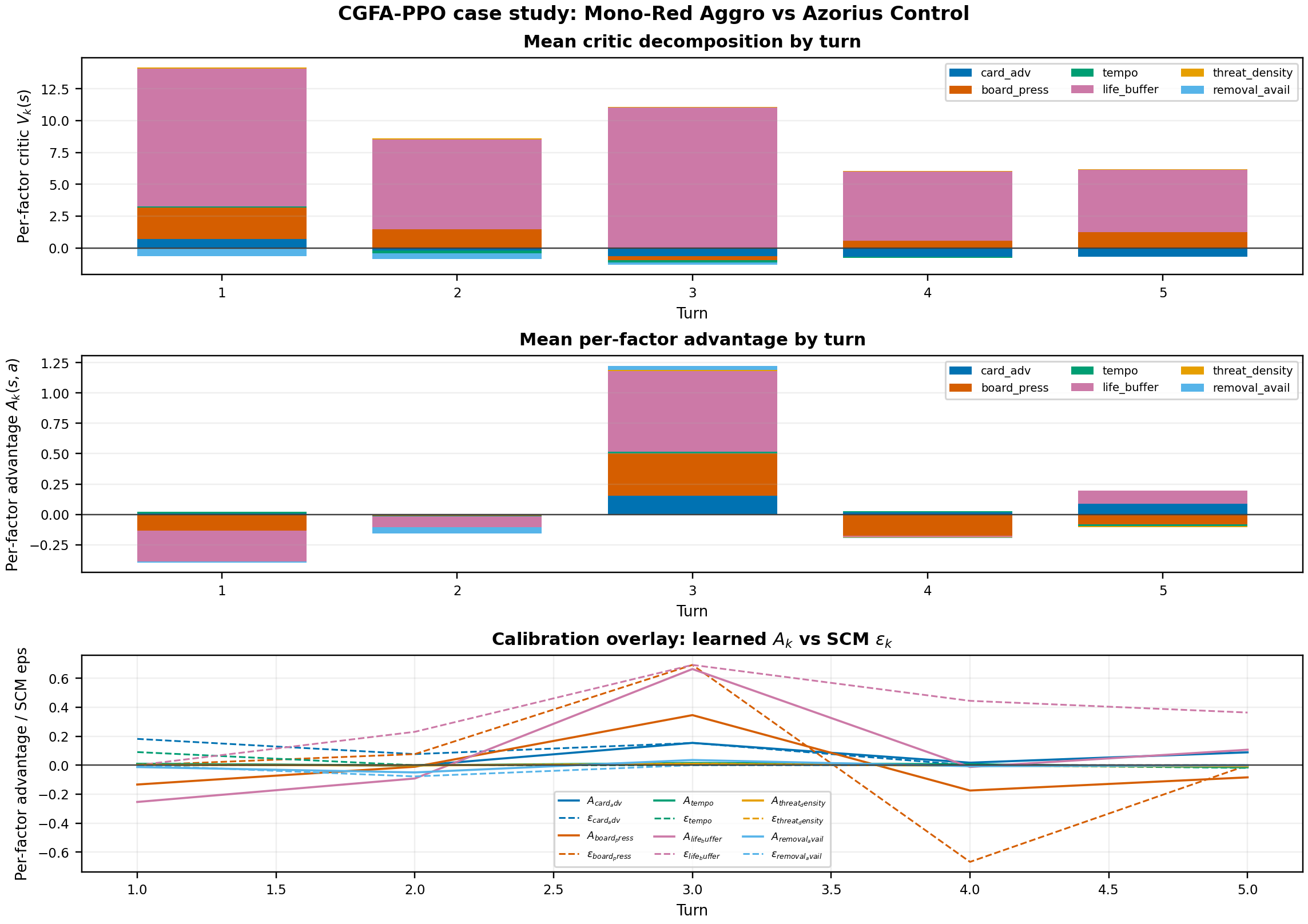}
\caption{CGFA-PPO case study on a single deterministic Mono-Red Aggro
player-deck episode against Azorius Control. (a) Mean stacked
per-factor critic $V_k(s)$ by turn, showing how each causal factor
contributes to the value estimate. (b) Mean stacked per-factor
advantage $A_k(s,a)$ by turn, showing how chosen actions are
credited across factors. (c) The intervention-calibration target
$\varepsilon_k(s,a)$ overlaid as a line trace on top of $A_k$ for all
factors, providing a turn-level visual check that the per-factor
heads track the SCM's intervention prediction.}
\label{fig:case_study}
\end{figure}

Figure~\ref{fig:headline_diagnostics} reports per-matchup win-rate
heatmaps complementing Figure~\ref{fig:headline_winrate} in the main
text; Figure~\ref{fig:case_study} shows a single deterministic
episode trace illustrating how multiple causal factors contribute to
individual decisions, complementing the calibration trajectory of
Figure~\ref{fig:calibration}.

\section{Reproducibility Manifest}
\label{app:repro}

Every training and evaluation run writes a manifest with the schema:
\begin{itemize}
    \item \textit{run}: training-pipeline schema version, run name,
    timestamp.
    \item \textit{git}: commit SHA, branch, dirty flag.
    \item \textit{python}: Python version, platform, and key library
    versions.
    \item \textit{lockfile}: SHA-256 digest of the dependency lockfile.
    \item \textit{invocation}: command arguments and host identifier.
    \item \textit{environment}: tracked subset of environment
    variables relevant to deterministic execution.
    \item \textit{config}: full hyperparameter dump (training
    config, env config, agent kwargs).
    \item \textit{seeds}: seed list and per-seed RNG fingerprints.
\end{itemize}
A run can therefore be re-derived from $\{\mathrm{git\,SHA},
\mathrm{lockfile}, \mathrm{seed}\}$ alone.

\section{Factor Schema}
\label{app:factors}

The K factors used by CGFA-PPO are exactly the parents of
$\mathrm{WinProb}$ in the SCM: card advantage, board pressure, tempo,
life buffer, threat density, and removal availability. At every step the
benchmark exposes factor values, factor rewards, and SCM-predicted
factor effects. Per-factor returns and advantages are stored alongside
the scalar reward, value, and log-probability tensors, with the same
truncation handling used for the scalar channel.

\section{Deck Archetypes}
\label{app:decks}

The five Standard 2025 archetypes shipped with the benchmark are
modelled on competitive tournament lists~\citep{wizards2024metagame,
mtggoldfish2024meta} and are kept fixed across runs.

\paragraph{Mono-Red Aggro (Tier 1).}
Aggressive curve: Mountain $\times 20$, Monastery Swiftspear $\times
4$, Heartfire Hero $\times 4$, Slickshot Show-Off $\times 4$, Phoenix
Chick $\times 4$, Play with Fire $\times 8$, Lightning Strike $\times
8$, Monstrous Rage $\times 8$.

\paragraph{Azorius Control (Tier 1).}
Counterspells, board wipes, late-game finisher: Plains $\times 6$,
Island $\times 6$, Adarkar Wastes $\times 4$, Restless Anchorage
$\times 4$, Deserted Beach $\times 4$, Haughty Djinn $\times 4$, No
More Lies $\times 8$, Make Disappear $\times 4$, Memory Deluge
$\times 8$, Sunfall $\times 8$, The Wandering Emperor $\times 4$.

\paragraph{Dimir Midrange (Tier 1).}
Disruption plus efficient threats: Swamp $\times 6$, Island $\times
5$, Underground River $\times 4$, Shipwreck Marsh $\times 4$,
Restless Reef $\times 4$, Preacher of the Schism $\times 4$,
Sheoldred, the Apocalypse $\times 4$, Faerie Mastermind $\times 4$,
Deep-Cavern Bat $\times 4$, Go for the Throat $\times 8$, Cut Down
$\times 4$, Duress $\times 4$, Make Disappear $\times 4$, Memory
Deluge $\times 1$.

\paragraph{Domain Ramp (Tier 2).}
Multicolor mana acceleration into domain payoffs: Forest $\times 8$,
Plains $\times 4$, Island $\times 4$, Swamp $\times 2$, Mountain
$\times 2$, Adarkar Wastes $\times 2$, Underground River $\times 2$,
Atraxa, Grand Unifier $\times 4$, Topiary Stomper $\times 8$,
Llanowar Elves $\times 4$, Leyline Binding $\times 4$, Up the
Beanstalk $\times 4$, Sunfall $\times 4$, Memory Deluge $\times 4$,
Go for the Throat $\times 4$.

\paragraph{Boros Convoke (Tier 2).}
Wide token strategy with convoke synergy: Plains $\times 10$,
Mountain $\times 6$, Battlefield Forge $\times 4$, Inspiring Vantage
$\times 4$, Warden of the Inner Sky $\times 4$, Resolute
Reinforcements $\times 4$, Knight-Errant of Eos $\times 4$,
Monastery Swiftspear $\times 4$, Phoenix Chick $\times 4$, Heartfire
Hero $\times 4$, Play with Fire $\times 4$, Lightning Strike
$\times 4$, Monstrous Rage $\times 4$.

\section{Compute Resources}
\label{app:compute}

All experiments are run on a single NVIDIA RTX 3090 GPU and an AMD
Ryzen 9 5900X (12 cores, 24 threads). Training $5 \times 10^{5}$
environment steps takes roughly 2 hours per seed; the full
six-variant ablation across 5 seeds and 5 archetypes is therefore on
the order of 150 GPU-hours. Using a TDP of 350 W and a US average
carbon intensity of 0.4 kg CO\textsubscript{2} per kWh, the carbon
footprint is approximately
$150 \times 0.35 \times 0.4 \approx 21$ kg CO\textsubscript{2}.

\section{Game Rules Summary}
\label{app:rules}

For readers unfamiliar with MTG, a brief summary of the rules used
by the benchmark; the full ruleset is published by Wizards of the
Coast~\citep{wotc2024mtgrules}.

\paragraph{Objective.}
Reduce the opponent's life total from 20 to 0.

\paragraph{Turn structure.}
Each turn consists of phases: Beginning (untap, upkeep, draw), Main
(play lands, cast spells), Combat (declare attackers, declare
blockers, damage), Second Main, End.

\paragraph{Mana.}
Lands produce mana, which is spent to cast spells. Each spell has a
mana cost (for example, 2R means 2 generic mana plus 1 red mana).
Players play at most one land per turn.

\paragraph{Card types.}
Creatures (attack and block), Instants (cast at any time), Sorceries
(main phase only), Enchantments and Artifacts (persistent effects).

\paragraph{Combat.}
Attacking creatures deal damage equal to their power. The defending
player may assign blockers. Unblocked creatures deal damage to the
defending player.

\paragraph{Mulligan.}
Before the game, players may mulligan (shuffle and redraw), drawing
one fewer card each time, capped at 3 mulligans (London mulligan).


\end{document}